\documentclass[a4paper,twoside]{article}

\usepackage{epsfig}
\usepackage{subcaption}
\usepackage{calc}
\usepackage{amssymb}
\usepackage{amstext}
\usepackage{amsmath}
\usepackage{amsthm}
\usepackage{multicol}
\usepackage{pslatex}
\usepackage{apalike}
\usepackage{algorithm2e}
\usepackage[bottom]{footmisc}
\usepackage{adjustbox}
\usepackage{booktabs} 
\usepackage{SCITEPRESS} 
\usepackage{color}
\usepackage{xcolor}

\begin{document}
\title{Partial Tensorized Transformers for Natural Language Processing}

\author{\authorname{Subhadra Vadlamannati\sup{1}, Ryan Solgi\sup{1}}
\affiliation{\sup{1}Mercer Island High School, 9100 SE 42nd St, Mercer Island, USA}
\affiliation{\sup{2}Department of Electrical and Computer Engineering, University of California Santa Barbara, Santa Barbara, USA}
\email{\{subhadra, solgi\}@ucsb.edu}
}


\abstract{The transformer architecture has revolutionized Natural Language Processing (NLP) and other machine-learning tasks, due to its unprecedented accuracy. However, their extensive memory and parameter requirements often hinder their practical applications.  In this work, we study the effect of tensor-train decomposition to improve the accuracy and compress transformer vision-language neural networks, namely BERT and ViT. We focus both on embedding-layer compression and partial tensorization of neural networks (PTNN) through an algorithmic approach. Our novel PTNN approach significantly improves the accuracy of existing models by up to 5\%, all without the need for post-training adjustments, breaking new ground in the field of tensor decomposition.
}

\onecolumn \maketitle \normalsize \setcounter{footnote}{0} \vfill

\section{\uppercase{Introduction}}
\label{sec:introduction}
The transformer architecture has been extremely influential in the field of Natural Language Processing (NLP) by achieving state-of-the-art performance in multiple downstream tasks \cite{Brown2020}. Specifically, transformer-based models have achieved state-of-the-art performance in machine translation, sentence representation, and language generation \cite{Brown2020}. However, due to the multi-head self-attention layers present in such architectures and the sheer number of parameters present in each layer (upwards of a million parameters during each forward pass of the model), transformers are computationally demanding in terms of both memory and computation time \cite{Gu2022}. 

Effective network compression for convolutional neural networks with fully-connected networks is still an open research question. Therefore, compressing Transformer networks is essential for their use on edge devices and in low-resource environments.
Tensor Decomposition works by representing higher-order tensors as a series of low-rank factors\cite{Oseledets2011}.

Tensor Decomposition (TD) works by representing higher-order tensors as a series of simpler, lower-order tensors via elementary operations \cite{Oseledets2011}. There are a variety of methods proposed to express tensors in a low-rank format factored form, including CANDECOMP PARAFAC (CP), tucker, and tensor-train decomposition \cite{BacciuMandic2020}. 

Several studies have explored the application of tensor decomposition methods to reduce the size of neural networks and improve their efficiency. For example, CP decomposition has been utilized to factorize weight matrices as the sum of rank-1 tensors, reducing the model’s memory requirements and speeding up inference time 8x \cite{MaldonadoChan2021}. 

Techniques like tensorized training can be applied to neural networks to achieve both faster model inference and also a large reduction in the number of parameters. More recently, tensorized training approaches have explored the possibility of updating the factored tensors of the weights of deep neural networks during the forward pass instead of the model weights directly \cite{BacciuMandic2020}. 

In the field of natural language processing, while tensor decomposition methods have been extensively studied for text-based neural networks, their application to multimodal neural networks has received limited attention. Multimodal neural networks such as ALIGN \cite{Jia2021} and VisualBERT \cite{Li2019}, have gained significant attention in recent years due to their remarkable ability to process both text and image data. These dual-encoder architectures require a larger number of parameters compared to traditional text-based models due to their need to capture information from many  modalities. 

While there has been existing research on applying traditional tensor decomposition methods to reduce the size of neural networks, their application on multimodal neural networks has been severely limited \cite{MaldonadoChan2021}. Existing methods have demonstrated that tensor decomposition methods can effectively reduce memory time and increase efficiency on neural networks while maintaining minimal accuracy loss (Zhong 2019; Bacciu and Mandic 2020).

In this paper, we focus on vision-language transformer-based models as prime candidates for models that require compression, as these models employ a dual-encoder architecture that requires double the amount of parameters traditional text-based models use. These models must encode both text and image data and then generate a latent space that can be employed on a variety of downstream tasks including visual question-answering, visual common-sense reasoning, natural language for visual reasoning, and region-to-phrase grounding. Vision-language tasks have a wide range of practical applications, including closed-captioning for the hard-of-hearing and screen reading capabilities \cite{Morris2018}.

This paper aims to advance the field of multimodal language processing by applying state-of-the-art tensor decomposition techniques to compress popular networks such as ViT and BERT while maintaining model accuracy. We additionally study the effect of tensor size and hyperparameter values during tensorized training. 

\begin{figure*}[]
    \centering
    \includegraphics[width=\textwidth]{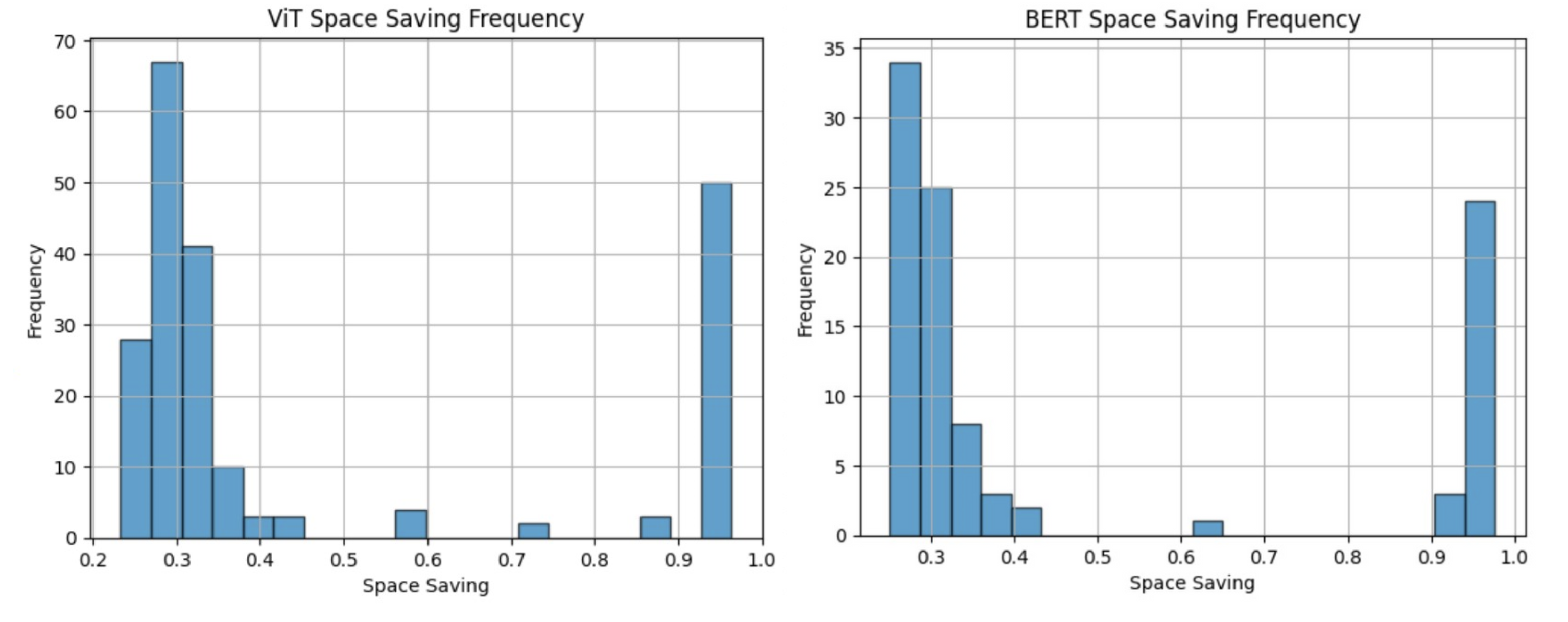}
    \caption{Bimodal Space-saving distribution for the two models.}
    \label{fig:babbage_asset_all}
\end{figure*}

\section{\uppercase{Methods}}
\label{sec:methods}
\subsection{Data}

We use the ImageNet dataset to train ViT. ImageNet is a collection of 80 million tiny images for image classification. Instead of using the entire dataset, we use the standard 1000 image test set of ImageNet known as CIFAR10 \cite{Krizhevsky2009} to evaluate the compressed model.

\subsection{Tensor-Train Decomposition}
\label{sec:tensortrain}
We use tensor-train (TT) decomposition throughout this paper to compress weight tensors. In the tensor train (TT) format \cite{Oseledets2011}, a $d$-way tensor $\textbf{$\mathcal{W}$} \in \mathbb{R}^{n_1\times .... \times n_d}$ is approximated with a set of $d$ cores $\bar{\textbf{$\mathcal{G}$}}=\{\textbf{$\mathcal{G}$}_1, \textbf{$\mathcal{G}$}_2, ..., \textbf{$\mathcal{G}$}_d\}$ where $\textbf{$\mathcal{G}$}_j \in \mathbb{R}^{r_{j-1}\times n_j \times r_{j}}$ , $r_j$'s for $j=1,...,d-1$ are the ranks, $r_0=r_d=1$, and each element of $\textbf{$\mathcal{Y}$}$ is approximated by the following equation:

\begin{align*} 
\label{eq:TT_format}
\hat{\textbf{{$\mathcal{W}$}}}[i_1,...,i_d] =\sum_{l_0,...,l_d} {\textbf{$\mathcal{G}$}_1[l_0,i_1,l_1]...\textbf{$\mathcal{G}$}_d[l_{d-1},i_d,l_d]}. 
\end{align*}

With a prescribed error tolerance ($\epsilon$), the fundamental components, denoted as \textbf{$\mathcal{G}_j$}, are obtained through a sequence of ($d-1$) consecutive Singular Value Decompositions (SVDs) performed on auxiliary matrices derived from the unfolding of the tensor \textbf{$\mathcal{Y}$} along various axes. This decomposition method, referred to as TT-SVD, is detailed in Algorithm \ref{TT-SVD}.

\begin{algorithm}[t]
  \caption{TT-SVD}
  \label{TT-SVD}
  \KwIn{$d$-way tensor \textbf{$\mathcal{Y}$}, error bound $\epsilon$.}
  \KwOut{$\bar{\textbf{$\mathcal{G}$}}=\{\textbf{$\mathcal{G}$}_1, \textbf{$\mathcal{G}$}_2, ..., \textbf{$\mathcal{G}$}_d\}$}
  $\sigma=\frac{\epsilon}{d-1}\Vert \textbf{$\mathcal{Y}$} \Vert_F$\\
  $r_0=1$, 
  $r_d=1$,  
 $\textbf{W}=$ reshape$(\mathcal{Y},(n_1,\frac{\vert \mathcal{Y} \vert}{n_1}))$\\
 \For{$j=1$ to $j=d-1$}{
 $\textbf{W}=$ reshape$(\textbf{W},(r_{j-1}n_j,\frac{\vert \textbf{W} \vert}{r_{j-1}n_j}))$\\
 Compute $\sigma$-truncated SVD: $\textbf{W}=\textbf{USV}^T+\textbf{E}$, where $\Vert \textbf{E} \Vert_F \leq \sigma$\\
 $r_j$ = the rank of matrix \textbf{W} based on $\sigma$-truncated SVD\\
 $\textbf{$\mathcal{G}$}_j=$ reshape$(\textbf{U},(r_{j-1},n_j,r_j))$\\
 $\textbf{W}=\textbf{SV}^T$\\
 }
 $\textbf{$\mathcal{G}$}_d=$ reshape$(\textbf{W},(r_{d-1},n_d,r_d))$\\
 Return $\bar{\textbf{$\mathcal{G}$}}=\{\textbf{$\mathcal{G}$}_1, \textbf{$\mathcal{G}$}_2, ..., \textbf{$\mathcal{G}$}_d\}$\\
\end{algorithm}

\subsection{Models and Pretraining}
\subsubsection{Models}
We perform an analysis on the effects of various models to encode both visual and textual data. For the text encoder, we focus on BERT \cite{Liu2019} which applies the bi-directional training of the Transformer to create in-context word embeddings. To test BERT on image classification, we pair it with EfficientNet for the visual encoding, which is consistent with the implementation in the dual-encoder network ALIGN \cite{ALIGN}. BERT has 12 layers, a hidden layer size of 768, and around 120~ million parameters. 

For the visual encoder we focus on ViT \cite{Dosovitskiy2021}. ViT is an image processor based on the Transformer architecture.

\subsubsection{Pre-Compression Training}

We fine-tune the image encoder ViT as it's trained on a much larger, more general dataset ImageNet. Additionally, this fine-tuning improves baseline model accuracy which ensures that our post-training adjustments truly improve model accuracy.
Specifically, we fine-tune ViT on the CIFAR10 train set with 50,000 images and hyperparameters as outlined in Section \ref{sec:hyperparameters}. Post fine-tuning, we achieve an accuracy of 97.89\%, a validation loss of 0.2564 and a training loss  of 0.4291.

\subsection{Embedding Layer Tensorization}
\label{sec:embeddinglayer}
Due to the relatively large size of the embedding layer, comprising around 28\% of BERT's total parameters, and around 40\% of ViT's total parameters, we first explore the effects of embedding layer tensorization. Following the aforementioned approach in \ref{sec:tensortrain}, we compress the embedding layers of the model and study the effects on model accuracy.

\subsection{Partially Tensorized Neural Network (PTNN)}
\label{sec:ptnn}
As the compression of the embedding layer caused the accuracy of the network to increase, as mentioned in \ref{sec:visualtransformer}, we continue exploring this accuracy improvement for the rest of the model using our method in \ref{sec:embeddinglayer}. In this case, we do not perform post-training adjustments (like retraining). Initially, we extract model weights from all relevant encoders. We compress every layer by initially transforming the weight's 2D matrix into a tensor of higher dimension with the same volume as the original matrix. Tensor-train decomposition is then used to decompose model weights with a given error bound. Lastly, we reshape the tensor back into the original dimensions of the 2D weight matrix. To perform this iteratively as described in Algorithm \ref{iterative-model-tensorization}, we re-initialize the model with the decomposed weights and perform the same process to compress the remaining model weights. 



\begin{algorithm}[!h]
 \caption{Iterative model tensorization.}
 \label{iterative-model-tensorization}
 \While{not at the end of the model}{
  compress layers $0$ to $n$ using TT decomp\;
  \eIf{model accuracy $\geq$ 5\% of the original accuracy}{
   proceed to layer $n+1$\;
   }{
   leave layer $n$ uncompressed\;
   move to layer $n+1$\;
  }
 }
\end{algorithm}

\begin{figure*}[ht]
    \centering  
    \includegraphics[scale=0.75]{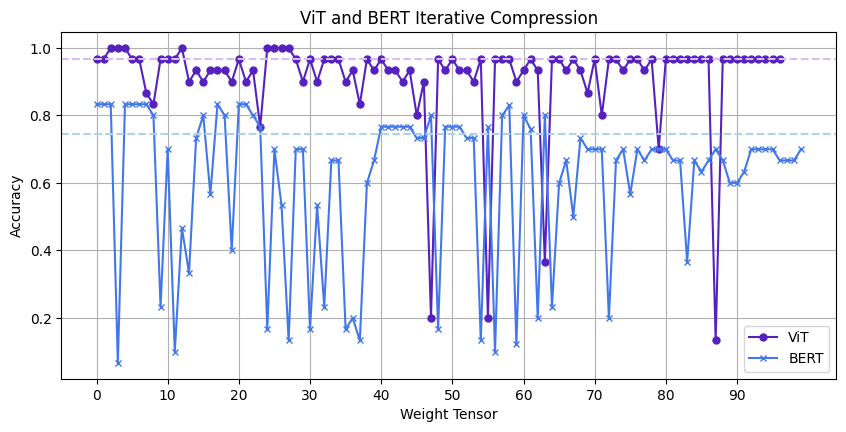}
    \caption{BERT and ViT model iterative compression with exclusion of low-accuracy layers. Lighter color corresponding dashed lines indicate baseline accuracy.}
    \label{fig:both_models_iterative}
\end{figure*}




\subsection{Decomposition Evaluation}

We evaluate the effectiveness of the decomposition using commonly known metrics like compression ratio and space saving, as described in Equation (1). 

We additionally evaluate the memory saved and time complexity of our decomposed models. For memory reduction, we simply multiply the percentage space saving with the total number of parameters in the non-decomposed matrix. Then, we divide this over the total number of parameters in the model.

For time complexity, we use a standard approach to find the O(n) based on the rank of the Tensor-Train decomposed matrices  as outlined in \cite{Oseledets2011}.

\section{\uppercase{Results}}
\label{sec:results}
We evaluate both models on both the train (50,000) and test (10,000) image split of the CIFAR10 dataset. The ViT model has 98 weight tensors while BERT contains 99 weight tensors. All layers are compressed with an epsilon of 0.5 and 3D projections are constructed with lowest norms.

\subsubsection{Embedding Layer Decomposition}
\label{sec:visualtransformer}

\begin{table}
\resizebox{\linewidth}{!}{%
\begin{tabular}{llll}
\toprule
\textbf{Model} & \textbf{Train} & \textbf{Test}  & \textbf{Space-saving} \\
\midrule
Decomposed ViT  & 0.986 & 0.966 & 0.267             \\
Base ViT                & 0.960  & 0.960  & NA                  \\
\midrule
Tensorized BERT (ALIGN) & 0.767 & 0.833 & 0.269              \\
Base BERT (ALIGN)       & 0.750 & 0.746 & NA       \\
\bottomrule
\end{tabular}
}
\caption{Embedding Layer Decomposition Results}
\label{tab:my-table}
\end{table}

The results of embedding layer decomposition are presented in Table \ref{tab:my-table}. Embedding layer decomposition results in an accuracy increase of around 2\% and still provides us with a noticeable parameter reduction. Even though we are not able to achieve high space-saving on these layers (i.e. only around 26\%), the success of this approach in maintaining or even improving accuracy leads us to apply this technique to compress more layers of the model.

As seen in \ref{sec:visualtransformer}, decomposing the embedding layer of BERT provides an accuracy gain over baseline comparisons. Following these promising results, we follow the methodology in Algorithm \ref{iterative-model-tensorization} to determine the amount of BERT that can be compressed whilst still maintaining overall accuracy.

\subsubsection{Iterative and Individual Decomposition}

We first individually compressed ViT's weight tensors, but did not saw decreases in model accuracy after the embedding layer (results found in \ref{sec:appendix}). Therefore, individual decomposition was not tested extensively on BERT due to computational limitations. Instead, we rather focused on the synergistic effects between layers of the model as mentioned in \ref{sec:ptnn}.

After compressing both models utilizing the process described in Algorithm \ref{iterative-model-tensorization}, the results are seen in \ref{fig:both_models_iterative}. In general, we see accuracy improvements even while compressing around half of the model's parameters (53\% in BERT and around 49\% in ViT). This is achieved all without post-training adjustments. Most notably, more layers in BERT tend to have an improved accuracy, however in ViT, compression of many layers does not change the baseline accuracy, while still resulting in a parameter reduction.

TT-decomposition can be an effective way to improve the accuracy of transformer-based models. While some layers vastly decrease the accuracy of the model, we can still achieve a high compression even when not including such layers in the overall decomposition.

\section{\uppercase{Conclusions}}
\label{sec:conclusion}
This paper introduces a novel approach to compressing transformer-based vision-language neural networks using TensorTrain decomposition. By applying this method to popular models like ViT and BERT, we achieved significant improvements in accuracy, up to 5\%, without the need for post-training adjustments. The iterative compression of model layers, coupled with retraining, enables us to preserve model accuracy while reducing up to 53\% of the model's parameters. In the future, we would like to generalize our approach to other dual encoder models and test our approach on other multimodal tasks like visual question answering and caption generation. This work represents a valuable advancement in the field of multimodal language processing and contributes to our broader goal of making transformer-based models more efficient and practical for real-world applications.

\section*{\uppercase{Acknowledgements}}
The first author would like to extend her gratitude to Dr. Lina Kim and the Research Mentorship Program Cohort for their support throughout the research process.

\bibliographystyle{apalike}
{\small
\bibliography{example}}

\section*{\uppercase{Appendix}}
\label{sec:appendix}
\subsection{ViT Fine-tuning Hyperparameters}
\label{sec:hyperparameters}
\begin{small}
\begin{verbatim}
Learning Rate: 5
Train/Evaluation Batch Size: 32
Optimizer: Adam, linear learning rate
Epoch(s): 1
\end{verbatim}
\end{small}

\subsection{Individual Compression}

\begin{figure}[h]
    \includegraphics[scale=0.3]{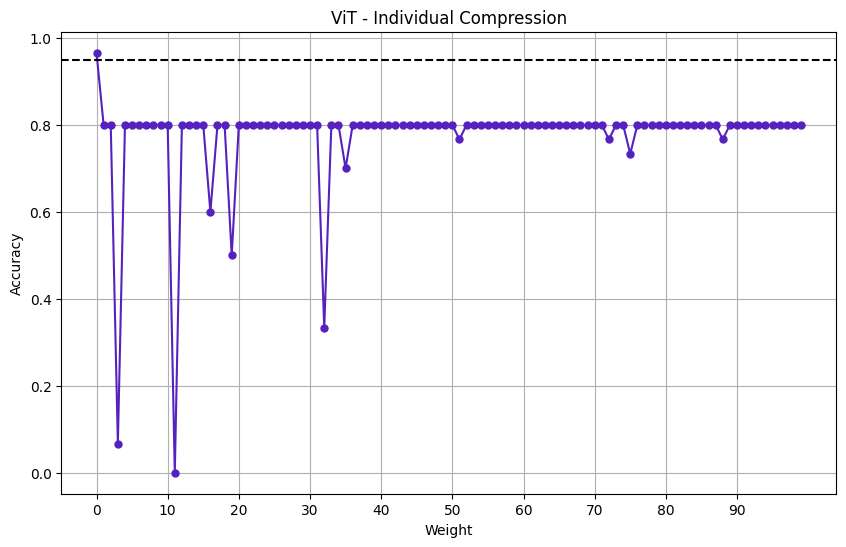}
    \caption{ViT Individual Compression}
    \label{fig:dataset_all_models}
\end{figure}


\end{document}